\pdfoutput=1
\documentclass[11pt]{article}

\usepackage[]{naacl2021}

\usepackage{times}
\usepackage{latexsym}
\usepackage{amsfonts}
\usepackage{graphicx}
\usepackage{amsmath}
\usepackage{xcolor}
\usepackage{multirow}
\usepackage{footnote}

\usepackage[T1]{fontenc}
\usepackage[utf8]{inputenc}

\usepackage{microtype}

\title{Improving Biomedical Pretrained Language Models with Knowledge}

\author{
Zheng Yuan$^{1}$\thanks{$\quad$Work done at Alibaba DAMO Academy.} \space\space\space
Yijia Liu$^{2}$ \space\space
Chuanqi Tan$^{2}$\thanks{$\quad$Corresponding author.} \space\space\space
Songfang Huang$^{2}$ \space\space
Fei Huang$^{2}$\\
$^{1}$Tsinghua University \space\space\space\space
$^{2}$Alibaba Group\\
\texttt{yuanz17@mails.tsinghua.edu.cn}\\
\texttt{\{yanshan.lyj,chuanqi.tcq,songfang.hsf,f.huang\}@alibaba-inc.com}
}

\begin{document}
\maketitle
\begin{abstract}
% The appearance of pretrained language models (LM) significantly improve the results in many natural language processing (NLP) tasks.
% Previous research show LMs trained on biomedical texts gain better performance on downstream tasks than LMs trained on Wikipedia or other general-domain corpora.
% In biomedical texts, entities play a vital role for understanding the implicit knowledge.
% However, these LMs trained in-domain does not utilize medical knowledge graphs which contain entities knowledge.
% Thus we propose KeBioLM: a knowledge graph enhanced biomedical LM.
% To fuse entity knowledge from the knowledge graph, KeBioLM extracts entities, and links entities to the Unified Medical Language System (UMLS).
% KeBioLM encodes sentences by a transformer encoder to text-only representations, and text-entity representations are generated by combining extracted entity information with another transformer encoder.
% We evaluate KeBioLM in named entity recognition and relation extraction tasks of BLURB benchmark, and achieve state-of-the-arts results.
% We further propose a cloze dataset based on UMLS relation triplets, and KeBioLM obtains better results than other biomedical LMs. 

Pretrained language models have shown success in many natural language processing tasks.
Many works explore incorporating knowledge into language models.
In the biomedical domain, experts have taken decades of effort on building large-scale knowledge bases.
For example, the Unified Medical Language System (UMLS) contains millions of entities with their synonyms and defines hundreds of relations among entities.
Leveraging this knowledge can benefit a variety of downstream tasks such as named entity recognition and relation extraction.
To this end, we propose KeBioLM, a biomedical pretrained language model that explicitly leverages knowledge from the UMLS knowledge bases.
Specifically, we extract entities from PubMed abstracts and link them to UMLS.
We then train a knowledge-aware language model that firstly applies a text-only encoding layer to learn entity representation and applies a text-entity fusion encoding to aggregate entity representation. Besides, we add two training objectives as entity detection and entity linking.
Experiments on the named entity recognition and relation extraction from the BLURB benchmark demonstrate the effectiveness of our approach. Further analysis on a collected probing dataset shows that our model has better ability to model medical knowledge.

\end{abstract}

\section{Introduction}
Large-scale pretrained language models (PLMs) are proved to be effective in many natural language processing (NLP) tasks \cite{peters-etal-2018-deep,devlin-etal-2019-bert}.
However, there are still many works that explore multiple strategies to improve the PLMs. Firstly, in specialized domains (i.e biomedical domain), many works demonstrate that using in-domain text (i.e. PubMed and MIMIC for biomedical domain) can further improve downstream tasks over general-domain PLMs \cite{lee2020biobert, peng-etal-2019-transfer, gu2020domain, shin-etal-2020-biomegatron, lewis-etal-2020-pretrained, beltagy-etal-2019-scibert, alsentzer-etal-2019-publicly}.
Secondly, unlike training language models (LMs) with unlabeled text, many works explore training the model with structural knowledge (i.e. triplets and facts) for better language understanding \cite{zhang-etal-2019-ernie,peters-etal-2019-knowledge,fevry-etal-2020-entities,wang2019kepler}. In this work, we propose to combine the above two strategies for a better \textbf{K}nowledge \textbf{e}nhanced \textbf{Bio}medical pretrained \textbf{L}anguage \textbf{M}odel (KeBioLM).

\begin{figure}[t]
\centering
\includegraphics[width=3in]{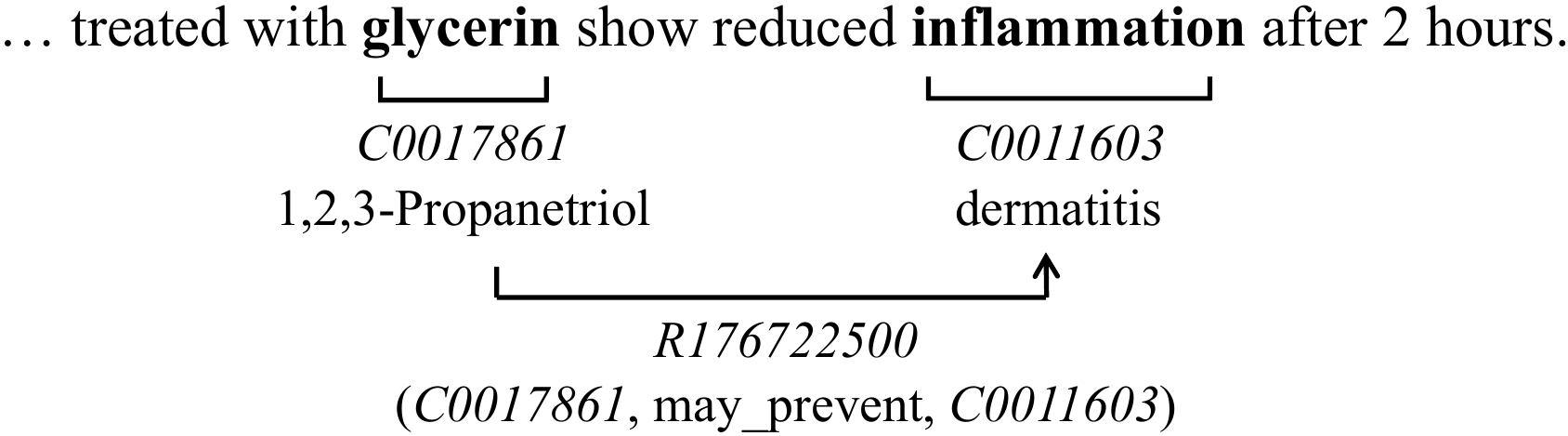}
\caption{An example of the biomedical sentence. Two entities ``glycerin'' and ``inflammation'' are linked to \textit{C0017861} (1,2,3-Propanetriol) and \textit{C0011603} (dermatitis) respectively with a relation triplet (\textit{C0017861}, \textit{may\_prevent}, \textit{C0011603}) in UMLS.}
\label{fig:example}
\end{figure}

As an applied discipline that needs a lot of facts and evidence, the biomedical and clinical fields have accumulated data and knowledge from a very early age \cite{ashburner2000gene, stearns2001snomed}. One of the most representative work is Unified Medical Language System (UMLS) \cite{bodenreider2004unified} that contains more than 4M entities with their synonyms and defines over 900 kinds of relations. Figure \ref{fig:example} shows an example. There are two entities ``glycerin'' and ``inflammation'' that can be linked to \textit{C0017861} (1,2,3-Propanetriol) and \textit{C0011603} (dermatitis) respectively with a \textit{may\_prevent} relation in UMLS. As the most important facts in biomedical text, entities and relations can provide information for better text understanding \cite{8377981, yuan2020coder}.

To this end, we propose to improve biomedical PLMs with explicit knowledge modeling. Firstly, we process the PubMed text to link entities to the knowledge base. We apply an entity recognition and linking tool ScispaCy \cite{neumann-etal-2019-scispacy} to annotate 660M entities in 3.5M documents.
Secondly, we implement a knowledge enhanced language model based on \citet{fevry-etal-2020-entities}, which performs a text-only encoding and a text-entity fusion encoding.
Text-only encoding is responsible for bridging text and entities. Text-entity fusion encoding fuses information from tokens and knowledge from entities. Finally, two objectives as entity extraction and linking are added to learn better entity representations.
To be noticed, we initialize the entity embeddings with TransE \cite{bordes2013translating}, which leverages not only entity but also relation information of the knowledge graph.

%we process the biomedical text to link the content to the knowledge base. We apply a entity recognition and linking tool ScispaCy to annotate entities to UMLS. Then, we train the KeBioLM on PubMed abstracts with masked language modeling, entity detection, and entity linking objectives following .

We conduct experiments on the named entity recognition (NER) and relation extraction (RE) tasks in the BLURB benchmark dataset. Results show that our KeBioLM outperforms the previous work with average scores of 87.1 and 81.2 on 5 NER datasets and 3 RE datasets respectively. Furthermore, our KeBioLM also achieves better performance in a probing task that requires models to fill the masked entity in UMLS triplets. 

We summary our contributions as follows\footnote{Our codes and model can be found at \url{https://github.com/GanjinZero/KeBioLM}.}:
\begin{itemize}
\item We propose KeBioLM, a biomedical pretrained language model that explicitly incorporates knowledge from UMLS.
\item We conduct experiments on 5 NER datasets and 3 RE datasets. Results demonstrate that our KeBioLM achieves the best performance on both NER and RE tasks.
\item We collect a cloze-style probing dataset from UMLS relation triplets. The probing results show that our KeBioLM absorbs more knowledge than other biomedical PLMs. 
\end{itemize}

\section{Related Work}
% In this section, we introduce the existing biomedical LMs and knowledge-enhanced LMs.

\subsection{Biomedical PLMs}
Models like ELMo \cite{peters-etal-2018-deep} and BERT \cite{devlin-etal-2019-bert} 
show 
the effectiveness 
of the paradigm of first pre-training an LM on the unlabeled text
then fine-tuning the model on the downstream NLP tasks.
However, 
direct application
of the LMs pre-trained on the encyclopedia and web text
usually fails
on the biomedical domain,
because of the distinctive terminologies and idioms.

The gap between general and biomedical domains
inspires the researchers to propose LMs specially tailored for the biomedical domain.
BioBERT \cite{lee2020biobert} is the most widely used biomedical PLM 
which is trained on PubMed abstracts and PMC articles. 
It outperforms vanilla BERT in named entity recognition, relation extraction, and question answering tasks.
\citet{jin2019probing} train BioELMo with PubMed abstracts, and find features extracted by BioELMo contain entity-type and relational information.
Different training corpora have been used for enhancing performance of sub-domain tasks.
ClinicalBERT \cite{alsentzer-etal-2019-publicly}, BlueBERT \cite{peng-etal-2019-transfer} and bio-lm \cite{lewis-etal-2020-pretrained} utilize clinical notes MIMIC to improve clinical-related downstream tasks.
SciBERT \cite{beltagy-etal-2019-scibert} uses papers from the biomedical and computer science domain as training corpora with a new vocabulary.
KeBioLM is trained on PubMed abstracts to adapt to PubMed-related downstream tasks.

To understand the factors in pretraining biomedical LMs, 
\citet{gu2020domain} study pretraining techniques systematically
and propose PubMedBERT pretrained from scratch with an in-domain vocabulary.
\citet{lewis-etal-2020-pretrained} also find using an in-domain vocabulary enhances the downstream performances.
This inspires us to utilize the in-domain vocabulary for KeBioLM.
% propose bio-lm which utilize structure and checkpoint of RoBERTa \cite{liu2019roberta}.

% \citet{peng-etal-2020-empirical} conduct multi-task learning of downstream tasks on existing LMs.

\subsection{Knowledge-enhanced LMs}
LMs like ELMo and BERT are trained to predict
correlation between tokens,
ignoring the meanings behind them.
To capture both the textual and conceptual information,
several knowledge-enhanced PLMs are proposed.
%Many researches are devoted to knowledge-enhanced LMs to link LMs and KGs.

Entities are used for bridging tokens and knowledge graphs.
\citet{zhang-etal-2019-ernie} align tokens and entities within sentences, and aggregate token and entity representations via two multi-head self-attentions.
KnowBert \cite{peters-etal-2019-knowledge} and Entity as Experts (EAE) \cite{fevry-etal-2020-entities} use the entity linker to perform entity disambiguation for candidate entity spans and enhance token representations using entity embeddings.
Inspired by entity-enhanced PLMs, we follow the model of EAE to inject biomedical knowledge into KeBioLM by performing entity detection and linking.

Relation triplets provide intrinsic knowledge between entity pairs.
KEPLER \cite{wang2019kepler} learns the knowledge embeddings through relation triplets while pretraining.
K-BERT \cite{Liu_Zhou_Zhao_Wang_Ju_Deng_Wang_2020} converts input sentences into sentence trees by relation triplets to infuse knowledge.

In the biomedical domain, \citet{he-etal-2020-infusing} inject disease knowledge to existing PLMs by predicting diseases names and aspects on Wikipedia passages.
\citet{michalopoulos2020umlsbert} use UMLS synonyms to supervise masked language modeling.
We propose KeBioLM to infuse various kinds of biomedical knowledge from UMLS including but not limited to diseases.

\section{Approach}

% In this paper, we assume to access a \textit{knowledge graph} $\mathcal{G}=\left<\mathcal{E}, \mathcal{R}\right>$,
In this paper, we assume to access an entity set $\mathcal{E}=\{e_1, ..., e_t\}$.
% where $\mathcal{E}$ is a set of entities
% and $\mathcal{R}$ is the relation that linking two entities.
For a sentence $\mathbf{x}=\{x_1, ..., x_n\}$,
we assume some spans $m=(x_i, ..., x_j)$ can be grounded to
one or more entities in $\mathcal{E}$.
We further assume the disjuncture of these spans.
In this paper, we use UMLS to set the entity set.

%To explicitly model the knowledge in UMLS, an instance is formulated as a sentence $\mathrm{S}$ with words $\{x_1, ..., x_N\}$ and mentions $\{m_1, ..., m_K\}$ with corresponding entities $\{e_1, ..., e_K\}$. $m_i$ is a mention span of $\{x_i,..., x_j\}$ and $e_i$ is the corresponding entity of $m_i$ in UMLS database.\footnote{Each entity is associated with a unique ID, which is called CUI.} 
%In the rest of this section,
%we will introduce data preparation, modeling architecture, training objectives, and implementation details. 

\subsection{Model Architecture}
\begin{figure*}[t]
\centering
\includegraphics[scale=0.45]{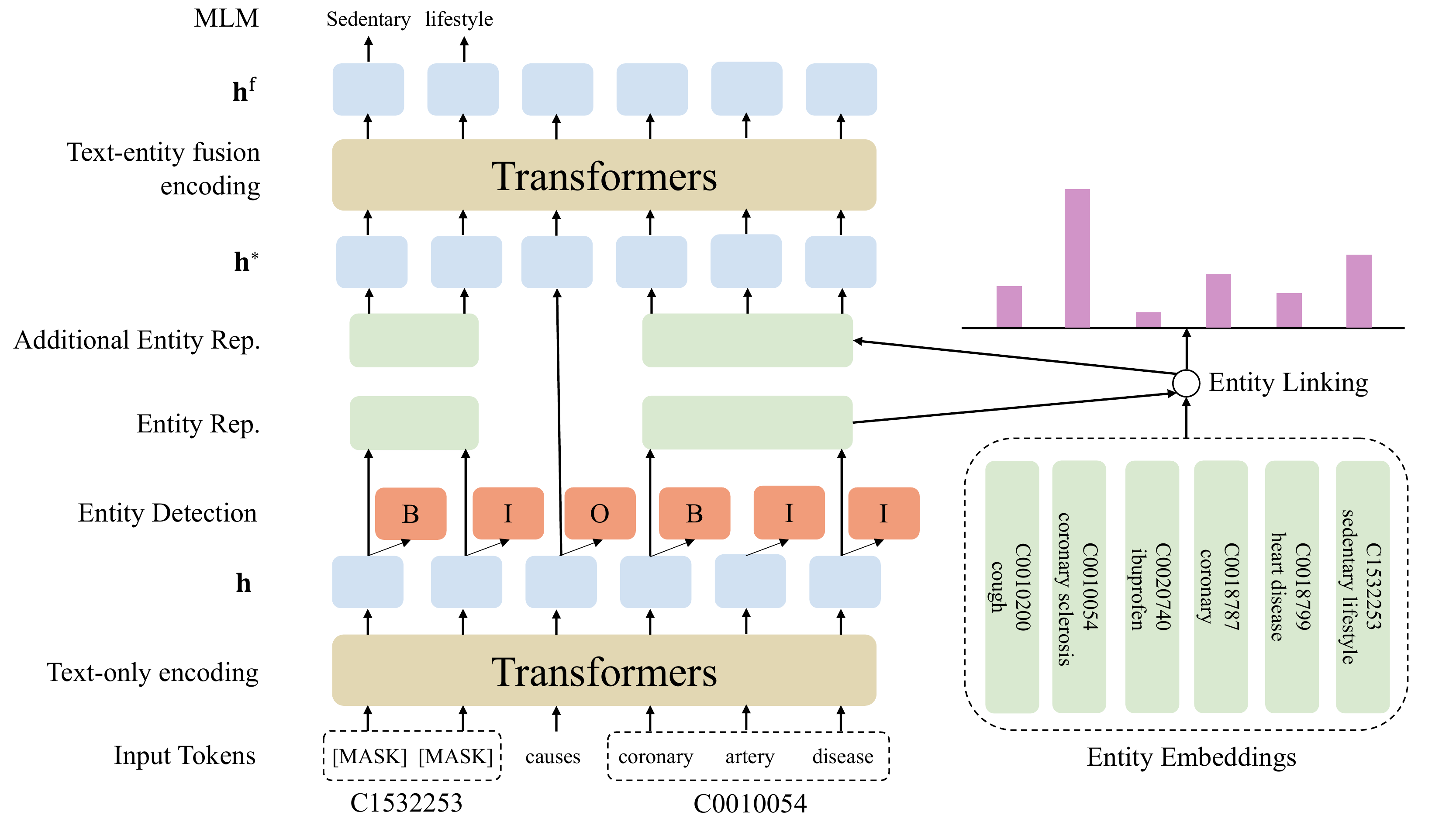}
\caption{The overall architecture of KeBioLM.}
\label{fig:model}
\end{figure*}

To explicitly model both the textual and conceptual information,
we follow \citet{fevry-etal-2020-entities} and 
use a multi-layer self-attention network to encode both the text and entities.
The model
can be viewed as
building the links between text and entities
in the lower layers and
fusing the text and entity representation in the upper layers.
The overall architecture is shown in Figure~\ref{fig:model}.
To be more specific,
we set the PubMedBERT \cite{gu2020domain}
as our backbone.
We split the layers of the backbone into two groups,
performing a text-only encoding and text-entity fusion encoding
respectively.

\paragraph{Text-only encoding.}
For the first group, which is closer to the input,
we extract the final hidden states
and perform a token-wise classification
to identify if the token is at the beginning, inside, or outside of a \textit{mention} (i.e., the BIO scheme).
The probabilities of the B/I/O label $\{l_i\}$ are written as:
% \begin{array}{rl}
\begin{align}
    \mathbf{h}_1, ... , \mathbf{h}_n &= {\rm Transformers}^{0}(x_1, ..., x_n) \\
     p(l_i \mid \mathbf{x} ) &= \textrm{softmax}(\mathbf{W}_l \mathbf{h}_i + \mathbf{b}_l)
\end{align}
% \end{array}
%We implement the knowledge enhanced model based on \citet{fevry-etal-2020-entities}. Figure~\ref{fig:model} shows the architecture.
%A sentence $x$ contains wordpieces $x_1, ..., x_N$ as input and contains $K$ entities $\{m_i = (e_{m_i}, s_i, t_i)\}_{i=1}^K$ where $e_{m_i} \in \mathcal{E}=\{e_1, ..., e_M\}$ is a pre-defined entity set, $s_i$ and $t_i$ indicate the start and end token index of entity $m_i$ in sentence $x$.
%Firstly, a series of Transformer Blocks are applied to encode the sentence. Here we use $l_0$ as the number of Transformer layers. Thus, a sentence is represented by $\mathbf{H}$ for hidden representations.
%An entity detection layer classifies each token into three entity position classes following the BIO scheme according to $\mathbf{H}$.
%In the BIO tagging scheme, B means the beginning of an entity, I means the inside of an entity, and O means outside of any entity.
%\begin{equation}
%    \hat{y}_i = {\rm softmax}(\mathbf{W}_dh_i + \mathbf{b}_d).
%\end{equation}
%BIO sequences can be decoded into start and end positions for each entity.
%For inference phase (fine-tuning phase), $\{m_i\}$ are not provided, thus we decode the predicted label of BIO sequences by $\{\hat{y_i}\}_{i=1}^N$ to extract entities $\{\hat{m}_i = (-, \hat{s}_i, \hat{t}_i)\}_{i=1}^{\hat{K}}$. 
%For the pretraining phase, we directly use $\{m_i\}$ from corpora.
After identifying the mention boundary,
we maintain a function $\mathcal{M}(i) \rightarrow \mathcal{E} \cup \{\textrm{NIL}\} $,
which returns the entity of the $i$-th token belongs.\footnote{NIL is returned when there is no entity being matched.}
We collect the mentions with a sentence $\mathbf{x}$.
For a mention $m=(s, t)$, where $s$ and $t$ represents the starting and ending indexes
of $m$,
we encode it
as the concatenation of hidden states of the boundary tokens $\mathbf{h}_m = [\mathbf{h}_{s}; \mathbf{h}_{t}]$.

For an entity $e_j \in \mathcal{E}$ in the KG,
we denote its entity embedding as $\mathbf{e}_j$.
For a mention $m$, we search
the $k$ nearest entities
of its projected representation $\mathbf{h}'_m = \mathbf{W}_m \mathbf{h}_m + \mathbf{b}_m$  
in the entity embedding space,
obtaining a set of entities $\mathcal{E}'$.
The normalized similarity between $\mathbf{h}'_m$ and $\mathbf{e}_j$
is calculated as
\begin{equation}
a_j = \frac{\exp(\mathbf{h}'_m \cdot \mathbf{e}_j)}
{\sum_{e_k\in \mathcal{E}'} \exp(\mathbf{h}'_m \cdot \mathbf{e}_k)}
\end{equation}
The additional entity representation $\mathbf{e}'_m$ of $m$
is calculated as a weighted sum
of the embeddings $\mathbf{e}'_m = \sum_{e_j\in \mathcal{E}'} a_j \cdot \mathbf{e}_j$.
%\begin{equation}
%   \mathbf{h}_{m} = \mathbf{W}_e[\mathbf{h}_{s}; \mathbf{h}_{t}] + \mathbf{b}_e.
%\end{equation}
%We assign $\mathbf{e}_i \in \mathbb{R}^d$ is the entity embedding for $e_i$ to memory entity knowledge.
%The similarities between entity $m_i$ in text and entities $e_j$ in $\mathcal{E}$ are calculated through dot product:
%\begin{equation}
%    S_{m_i,j} = \mathbf{h}_{m_i} \cdot \mathbf{e}_j.
%\end{equation}
%We use a weighted sum of top-k similar entities of $m_i$ in $\mathcal{E}$ as the entity representation $\mathbf{E}_{m_i}$ for $m_i$.
%\begin{equation}
%    A_i = {\rm argmax}_{A\subset \mathcal{E}, \|A\|=k}\sum_{j\in A}S_{m_i,j},
%\end{equation}
%\begin{equation}
%    \mathbf{E}_{m_i} = \sum_{j\in{A_i}}\frac{\exp(S_{m_i,j})\mathbf{e}_j}{\sum_{j\in{A_i}}\exp(S_{m_i,j})}.
%\end{equation}

\paragraph{Text-entity fusion encoding.}
After getting the mentions
and entities,
we fuse the entity embeddings with the text embedding by summation.
For the $i$-th token,
the entity-enhanced embedding is calculated
as:
\begin{equation}
\mathbf{h}^*_i = 
\begin{cases}
     \mathbf{h}_i + \left(\mathbf{W}_e \mathbf{e}'_m + \mathbf{b}_e\right), & \exists m, \mathcal{M}(i) = m,  \\
     \mathbf{h}_i, & \textrm{otherwise.}
\end{cases}
\end{equation}
$\mathcal{M}(i) = m$ represents the $i$-th token belong to entity $e_m$.
The sequence of $\mathbf{h}^*_1, ..., \mathbf{h}^*_n$
is then fed into the second group of transformer layers
to generate text-entity representations.
The final hidden states $\mathbf{h}^f_i$ are calculated as:
% both the text and entity.
\begin{equation}
    \mathbf{h}^f_1, ... , \mathbf{h}^f_n = {\rm Transformers}^{1}(\mathbf{h}^*_1, ..., \mathbf{h}^*_n)
\end{equation}
%To get an entity-aware token representation, we add a linear transformation of the entity representation $\mathbf{E}_{m_i}$ on all tokens of $m_i$.
%If $s_i \leq j \leq t_i$, we have:
%\begin{equation}
%    \mathbf{h}_j^e = \mathbf{h}_j + (\mathbf{W}_b\mathbf{E}_{m_i} + \mathbf{b}_b).
%\end{equation}
%Otherwise, $\mathbf{h}_j^e = \mathbf{h}_j$.
%$\mathbf{H}^e = (\mathbf{h}_1^e, ... , \mathbf{h}_N^e)$ contain the entity knowledge via entity representation $\mathbf{E}_{m_i}$.

%We apply a LayerNorm on $\mathbf{H}^e$ and feed it into another $l_1$ transformer layers to encode the final hidden representations $\mathbf{H}^f$.
%\begin{equation}
%    \mathbf{H}^f = {\rm Transformer}^{l_1}({\rm LayerNorm}(\mathbf{H}^e)).
%\end{equation}

\subsection{Pretraining Tasks}
We have three pretraining tasks for KeBioLM.
Masked language modeling is a cloze-style task for predicting masked tokens.
Since the entities are the main focus of our model, we add two tasks as entity detection and linking respectively following \citet{fevry-etal-2020-entities}.
Finally, we jointly minimize the following loss:
\begin{equation}
    \mathcal{L} = \mathcal{L}_{MLM} + \mathcal{L}_{ED} + \mathcal{L}_{EL}
\end{equation}

\paragraph{Masked Language Modeling}
Like BERT and other LMs, we predict the masked tokens $\{x_i\}$ in inputs using the final hidden representations $\{\mathbf{h}^f_i\}$. The loss $\mathcal{L}_{MLM}$ is calculated based on the cross-entropy of masked and predicted tokens:
\begin{align}
    p_{M}(x_i \mid \mathbf{x}) &= {\rm softmax}(\mathbf{W}_m\mathbf{h}^f_i + \mathbf{b}_m) \\
    \mathcal{L}_{MLM} &= \sum-\log p_{M}(x_i \mid \mathbf{x})
\end{align}
% Inspired by the success of whole word masking in training masked language models \cite{devlin-etal-2019-bert,cui2019pre}, we extend this technique to whole entity masking.
Whole word masking is successful in training masked language models \cite{devlin-etal-2019-bert,cui2019pre}.
In the biomedical domain, entities are the semantic units of texts.
Therefore, we extend this technique to whole entity masking.
We mask all tokens within a word or entity span.
KeBioLM replaces 12\% of tokens to \textit{[MASK]} and 1.5\% tokens to random tokens.
This is more difficult for models to recover tokens, which leads to learning better entity representations.

\paragraph{Entity Detection}
Entity detection is an important task in biomedical NLP to link the tokens to entities. Thus, We add an entity detection loss by calculating the cross-entropy for BIO labels:
\begin{equation}
    \mathcal{L}_{ED} = \sum_{i=1}^n -\log p(l_i\mid \mathbf{x})
\end{equation}
% KeBioLM can serve as a ready-to-work NER model which extracts biomedical entities in multiple categories.

\paragraph{Entity Linking}
One medical entity in different names linking to the same index permits the model to learn better text-entity representations.
To link mention $\{m\}$ in texts with entities $\{e\}$ in entity set $\mathcal{E}$, we calculate the cross-entropy loss using similarities between $\{\mathbf{h}'_m\}$ and entities in $\mathcal{E}$:
\begin{equation}
    \mathcal{L}_{EL} = \sum -\log\frac{\exp(\mathbf{h}'_m\cdot \mathbf{e})}{\sum_{e_j\in\mathcal{E}}\exp(\mathbf{h}'_m\cdot \mathbf{e}_j)}
\end{equation}
% For the inference phase, entity linking is calculated on $\{\hat{m}_i\}$.

\subsection{Data Creation}

Given a sentence $\mathrm{S}$ from PubMed content, we need to recognize entities and link them to the UMLS knowledge base. We use ScispaCy \cite{neumann-etal-2019-scispacy}, a robust biomedical NER and entity linking model, to annotate the sentence. Unlike previous work \cite{vashishth2020medtype} that only retains recognized entities in a subset of Medical Subject Headings (MeSH) \cite{lipscomb2000medical}, we relax the restriction to annotate all entities to UMLS 2020 AA release \footnote{\url{https://www.nlm.nih.gov/research/umls/licensedcontent/umlsarchives04.html\#2020AA}} whose linking scores are higher than a threshold of 0.85.

% We leverage the PubMedDS dataset \cite{vashishth2020medtype} for KeBioLM pretraining.
% PubMedDS extract and link entities to the Unified Medical Language System (UMLS) \cite{bodenreider2004unified} Concept Unique Identifier (CUI) via distant supervision on PubMed abstracts.
% However, PubMedDS only retain entities in a subset of Medical Subject Headings which results in a low recall.
% We relax this restriction to recognize medical entities from PubMedDS in the whole UMLS for pretraining.

% To relabel PubMedDS entities, we use ScispaCy \cite{neumann-etal-2019-scispacy} which is a robust biomedical NER and entity linking model.
% \texttt{en\_core\_sci\_lg} model is applied for NER.
% ScispaCy links entities to a subset of UMLS 2017 AA release based on character 3-gram TF-IDF score.
% For each entity, we only retain at most one linked CUI over a score threshold of 0.85.

\section{Experiments}
In this section, we first introduce the pretraining details of KeBioLM. 
Then we introduce the BLURB datasets for evaluating our approach.
Finally, we introduce a probing dataset based on UMLS triplets for evaluating knowledge modeling.
% \subsection{Training Data}
% After relabeling by ScispaCy, we acquire 477,036 linked CUIs though there are only 32,742 CUIs in the original version of PubMedDS.
% Table~\ref{tab:pubmedds} lists the summary of the original and relabeled PubMedDS dataset.

% \begin{table}[ht]
% \centering
% \begin{tabular}{cccc}
% \hline
% Dataset&\#Docs&\#Entities&\#CUIs\\
% \hline
% % PubMedDS& 3,471,137&5,238,053&32,742\\
% % relabeled& 3,471,137&659,767,076&477,036\\
% PubMedDS& 3.5M & 5.2M & 32.7K\\
% + relabeled& 3.5M &660M & 477K \\
% \hline
% \end{tabular}
% \caption{The summary of original and relabeled PubMedDS dataset. ScispaCy is used for constructing relabeled version.}
% \label{tab:pubmedds}
% \end{table}

% &$\#$Train&$\#$Dev&$\#$Test&$\#$Ments&$\#$Ments in UMLS&$\#$Ments in KeBioLM\\
\begin{table*}[t]
\centering
\begin{tabular}{lccc|ccc}
\hline
&\multirow{2}*{$\#$Train}
&\multirow{2}*{$\#$Dev}
&\multirow{2}*{$\#$Test}
&\multirow{2}*{$\#$Ments}
&\multirow{2}*{\shortstack{$\#$Ments \\(UMLS)}}
&\multirow{2}*{\shortstack{$\#$Ments \\(KeBioLM)}}
\\
&&&&&&\\
\hline
BC5chem&5,203&5,347&5,385&15,935&10,373&8,993\\
BC5dis&4,182&4,244&4,424&12,850&8,846&3,878\\
NCBI&5,137&787&960&6,884&1,985&1,091\\
BC2GM&15,197&3,061&6,325&24,583&2,808&2,423\\
JNLPBA&46,750&4,551&8,662&59,963&6,099&5,233\\
\hline
ChemProt&18,035&11,268&15,745&39,022&13,106&10,772\\
DDI&25,296&2,496&5,716&15,738&10,429&9,212\\
GAD&4,261&535&534&-&-&-\\
\hline
\end{tabular}
\caption{The training instances (mentions for NER tasks and sentences with two entities for RE tasks) and the mention counts of NER and RE datasets preprocessed in BLURB benchmark respectively. The mention counts overlapping with UMLS 2020 AA release and KeBioLM are also listed.
For the GAD dataset, annotated mentions do not appear in the BLURB preprocessed version.}
%We report the sizes of training, development and testing set.}
\label{tab:blurb}
\end{table*}

\subsection{Pretraining Details}
We use ScispaCy to acquire 477K CUIs and 660M entities among 3.5M PubMed documents\footnote{The count of documents in PubMedDS is based on \url{https://arxiv.org/pdf/2005.00460v1.pdf}.} from PubMedDS dataset \cite{vashishth2020medtype} as training corpora.

We initialize entity embeddings by TransE \cite{bordes2013translating} which learns embeddings from relation triplets.
Relation triplets come from UMLS 2020 AA release.
% contain 477K CUIs extracted by ScispaCy .
We train TransE with the L2-norm distance function and set embedding dim to 100.
Adam \cite{kingma2014adam} is used as the optimizer with a learning rate of 1e-3, batch size of 2048, and train epoch of 30.
Entity embeddings add 45.5M parameters to KeBioLM.

The parameters of transformers in KeBioLM are initialized from the checkpoint of PubMedBERT.
We also use the vocabulary from PubMedBERT.
% Transformers and entity embeddings are both trainable.
AdamW \cite{loshchilov2017decoupled} is used as the optimizer for KeBioLM with 10,000 steps warmup and linear decay.
% We use $l_0=8$ and $l_1=4$ for transformer layers.
We use an 8-layer transformer for text-only encoding and a 4-layer transformer for text-entity fusion encoding.
We set the learning rate to 5e-5, batch size to 512, max sequence length to 512, and training epochs to 2.
For each input sequence, we limit the max entities count to 50 and the excessive entities will be truncated.
To generate entity representation $\mathbf{e}'_m$, the most $k=100$ similar entities are used.
We train our model with 8 NVIDIA 16GB V100 GPUs.

\subsection{Datasets}
In this section, we evaluate KeBioLM on NER tasks and RE tasks of the BLURB benchmark\footnote{\url{https://microsoft.github.io/BLURB/}} \cite{gu2020domain}.
% To notice, our pretraining corpora are labeled by ScispaCy.
% ScispaCy is trained on MedMentions dataset \cite{mohan2019medmentions} and is different from datasets in BLURB.
For all tasks, we use the pre-processed version from BLURB.
We measure the NER and RE datasets in terms of F1-score.
Table~\ref{tab:blurb} shows the counts of training instances in BLURB datasets (i.e., annotated mentions for NER datasets and sentences with two mentions for RE datasets).
We also report the count of annotated mentions overlapping with the UMLS 2020 release and KeBioLM in each dataset.
The percentage of mentions overlapping with KeBioLM ranges from 8.7\% (NCBI-disease) to 58.5\% (DDI) which indicates that KeBioLM learns entity knowledge related to downstream tasks.

\subsubsection{Named Entity Recognition}
\paragraph{BC5-chem \& BC5-disease} \cite{li2016biocreative} contain 1500 PubMed abstracts for extracting chemical and disease entities respectively.

\paragraph{NCBI-disease} \cite{dougan2014ncbi} includes 793 PubMed abstracts to detect disease entities.

\paragraph{BC2GM} \cite{smith2008overview} contains 20K PubMed sentences to extract gene entities.

\paragraph{JNLPBA} \cite{collier-kim-2004-introduction} includes 2,000 PubMed abstracts to identify molecular biology-related entities.
We ignore entity types in JNLPBA following \citet{gu2020domain}.

\subsubsection{Relation Extraction}

\paragraph{ChemProt} \cite{krallinger2017overview} classifies the relation between chemicals and proteins within sentences from PubMed abstracts.
Sentences are classified into 6 classes including a negative class.

\paragraph{DDI} \cite{herrero2013ddi} is a RE dataset with sentence-level drug-drug relation on PubMed abstracts.
There are four classes for relation: advice, effect, mechanism, and false.

\paragraph{GAD} \cite{bravo2015extraction} is a gene-disease relation binary classification dataset collected from PubMed sentences.

\begin{table*}[ht]
\centering
\begin{tabular}{lccccccc|l}
\hline
&\multirow{2}*{\shortstack{Bio-\\BERT}}
&\multirow{2}*{\shortstack{Sci-\\BERT}}
&\multirow{2}*{\shortstack{Clinical-\\BERT}}
&\multirow{2}*{\shortstack{Blue-\\BERT}}
&\multirow{2}*{\shortstack{disease-\\BERT}}
&\multirow{2}*{\shortstack{bio-\\lm†}}
&\multirow{2}*{\shortstack{PubMed-\\BERT}}
&\multirow{2}*{\shortstack{KeBio-\\LM}}
\\
&&&&&&&&\\
\hline
BC5chem&92.9&92.5&90.8&91.2&-&92.9&\textbf{93.3}&$\textbf{93.3}_{\pm 0.2}$\\
BC5dis&84.7&84.5&83.0&83.7&\textbf{86.5}&83.8&85.6&$86.1_{\pm 0.3}*$\\
NCBI&\textbf{89.1}&88.1&88.3&88.0&87.1&87.7&87.8&$\textbf{89.1}_{\pm 0.3}*$\\
BC2GM&83.8&83.4&81.7&81.9&-&87.0&84.5&$\textbf{85.1}_{\pm 1.6}$\\
JNLPBA&79.4&79.5&78.6&78.7&-&80.6&80.1&$\textbf{82.0}_{\pm 0.2}*$\\
\hline
NER&86.0&85.6&84.5&84.7&-&86.4&86.3&$\textbf{87.1}_{\pm 0.3}*$\\
\hline
ChemProt&76.1&75.2&72.0&71.5&-&75.4&77.2&$\textbf{77.5}_{\pm 0.3}*$\\
DDI&80.9&81.1&78.2&77.8&-&81.0&\textbf{82.4}&$81.9_{\pm 0.8}$\\
GAD&80.9&80.9&78.4&77.2&-&82.2&82.3&$\textbf{84.3}_{\pm 1.0}*$\\
\hline
RE&79.3&79.1&76.2&75.5&-&79.5&80.6&$\textbf{81.2}_{\pm 0.5}*$\\
\hline
\end{tabular}

\caption{F1-scores on NER and RE tasks in BLURB benchmark.
Standard deviations of KeBioLM are reported across five runs.
Results of diseaseBERT-biobert and bio-lm come from their corresponded papers.
Others are copied from BLURB.
% Some results of bio-lm \cite{lewis-etal-2020-pretrained} are reported with different metrics according to their paper.\textcolor{red}{micro/macro f1}
* indicates that $p \leq 0.05$ of one-sample t-test which compares whether the mean performance of KeBioLM is better than PubMedBERT.
† Bio-lm applies different metrics with BLURB (micro F1 v.s. macro F1). Thus, we just list its results but do not directly compare with them.
}
\label{tab:main}
\end{table*}

\subsection{Fine-tuning Details}
\paragraph{NER} We follow \citet{gu2020domain} to formulate NER tasks as sequential labeling tasks with the BIO tagging scheme and ignore the entity types in NER datasets.
We classify labels of tokens by a linear layer on top of the hidden representations.

\paragraph{RE} We replace the entity mentions in RE datasets with entity indicators like @DISEASE\$ or @GENE\$ to avoid models classifying relations by memorizing entity names.
We add these entity indicators into the vocabulary of LMs.
We concatenate the representation of two concerned entities and feed it into a linear layer for relation classification.

\paragraph{Parameters} We adopt AdamW as the optimizer with a 10\% steps linear warmup and a linear decay.
We search the hyperparameters of learning rate among 1e-5, 3e-5, and 5e-5.
We fine-tune the model for 60 epochs.
We evaluate the model at the end of each epoch and choose the best model according to the evaluation score on the development set.
We set batch size as 16 when fine-tuning.
The maximal input lengths are 512 for all NER datasets.
We truncate ChemProt and DDI to 256 tokens, and GAD to 128 tokens.
To perform a fair comparison, we fine-tune our model with 5 different seeds and report the average score.

\subsection{Results}
We compare KeBioLM with following base-size biomedical PLMs on the above-mentioned datasets: 
BioBERT \cite{lee2020biobert},
SciBERT \cite{beltagy-etal-2019-scibert}, 
ClinicalBERT \cite{alsentzer-etal-2019-publicly},
BlueBERT \cite{peng-etal-2019-transfer},
% MT-BioBERT-Fine-Tune \cite{peng-etal-2020-empirical} \textcolor{red}{Do I need to compare with this?},
bio-lm \cite{lewis-etal-2020-pretrained},
diseaseBERT \cite{he-etal-2020-infusing},
and PubMedBERT \cite{gu2020domain} \footnote{We use BioBERT v1.1, SciBERT-scivocab-uncased, Bio-ClinicalBERT, BlueBERT-pubmed-mimic, bio-lm(RoBERTa-base-PM-M3-Voc), diseaseBERT-biobert and PubMedBERT-abstract versions for comparison.}.

Table~\ref{tab:main} shows the main results on NER and RE datasets of the BLURB benchmark.
In addition, we report the average scores for NER and RE tasks respectively.
KeBioLM achieves state-of-the-art performance for NER and RE tasks.
Compared with the strong baseline BioBERT, KeBioLM shows stable improvements in NER and RE datasets (+1.1 in NER, +1.9 in RE).
% Compared with our baseline model PubMedBERT, KeBioLM achieves better results in almost all datasets (equal for BC5chem, -0.5 for DDI) and achieves better average scores in both two tasks (+0.8 in NER, +0.6 in RE).
Compared with our baseline model PubMedBERT, KeBioLM performs significantly better in BC5dis, NCBI, JNLPBA, ChemProt, and GAD ($p \leq 0.05$ based on one-sample t-test) and achieves better average scores (+0.8 in NER, +0.6 in RE).
% KeBioLM averagely outperforms other LMs on NER tasks (+0.6\% compared to bio-lm) and RE tasks (+0.7\% compared to PubMedBERT) and achieves best results across 6 of 8 datasets.
DiseaseBERT is a model carefully designed for predicting disease names and aspects, which leads to better performance in the BC5dis dataset (+0.4). They only report the promising results in disease-related tasks, however, our model obtains consistent promising performances across all kinds of biomedical tasks.
% Bio-lm report better results on BC2GM dataset (+2.3\% compared to us).
% The results for same model on BC2GM in bio-lm paper are constantly higher than results in BLURB benchmark (BioBERT 85.6 vs 83.8, SciBERT 85.7 vs 83.4, ClinicalBERT 83.9 vs 81.7).
In the BC2GM dataset, KeBioLM outperforms our baseline model PubMedBERT and other PLMs except for bio-lm, and the standard deviation of the BC2GM task is evidently larger than other tasks.
Another exception is the DDI dataset, we observe a slight performance degradation compared to PubMedBERT (-0.5).
The average performances demonstrate that fusing entity knowledge into the LM boosts the performances across the board.

\subsection{Ablation Test}
We conduct ablation tests to validate the effectiveness of each part in KeBioLM.
We pretrain the model with the following settings and reuse the same parameters described above:
(a) Remove whole entity masking and retain whole word masking while pretraining (-wem);
(b) Initialize entity embeddings randomly (+rand);
(c) Initialize entity embeddings by TransE and freeze the entity embeddings while pretraining (+frz).

% We pretrain and evaluate these settings.
In Table~\ref{tab:ablation}, we observe the following results.
Firstly, comparing KeBioLM with setting (a) shows that whole entity masking boosting the performances consistently in all datasets (+0.5 in NER, +0.9 in RE).
Secondly, comparing KeBioLM with setting (b) indicates initializing the entity embeddings randomly degrades performances in NER tasks and RE tasks (-0.4 in NER, -1.2 in RE).
Entity embeddings initialized by TransE utilize relation knowledge in UMLS and enhance the results. 
Thirdly, freezing the entity embeddings in setting (c) reduces the performances on all datasets compared to KeBioLM except BC2GM (-0.4 in NER, -1.1 in RE).
This indicates that updating entity embedding while pretraining helps KeBioLM to have better text-entity representations, and this leads to better downstream performances.

\begin{table}
\centering
\begin{tabular}{lc|ccc}
\hline
&\multirow{2}*{\shortstack{KeBio-\\LM}}
&\multirow{2}*{\shortstack{-wem}}
&\multirow{2}*{\shortstack{+rand}}
&\multirow{2}*{\shortstack{+frz}}
\\
&&&&\\
\hline
BC5chem&\textbf{93.3}&92.8&92.8&92.3\\
BC5dis&\textbf{86.1}&85.9&85.5&85.5\\
NCBI&\textbf{89.1}&88.4&88.8&88.3\\
BC2GM&85.1&84.5&84.5&\textbf{85.7}\\
JNLPBA&\textbf{82.0}&81.5&81.9&81.8\\
\hline
NER&\textbf{87.1}&86.6&86.7&86.7\\
\hline
ChemProt&\textbf{77.5}&77.3&76.3&76.8\\
DDI&\textbf{81.9}&80.6&81.4&80.7\\
GAD&\textbf{84.3}&83.1&82.3&82.8\\
\hline
RE&\textbf{81.2}&80.3&80.0&80.1\\
% \hline
% Type 1&14.01&14.06\\
% Type 2&1.48&1.64\\
% Type 1&6.23&6.21&\textbf{6.33}&5.32\\
% % Type 2&\textbf{0.59}&0.55&\textbf{0.59}&0.36\\
% \hline
% Overall&3.26&3.45\\
% Overall&\textbf{1.44}&1.35&1.41&1.10\\
\hline
\end{tabular}
\caption{Ablation studies for KeBioLM architecture on the BLURB benchmark. We use -wem, +rand and +frz to represent pretraining setting (a), (b) and (c), respectively.}
\label{tab:ablation}
\end{table}

To evaluate how the count of transformer layers affects our model, we pretrain KeBioLM with the different number of layers.
For the convenience of notation, denote $l_0$ is the layer count of text-only encoding and $l_1$ is the layer count of text-entity fusion encoding.
We have the following settings:
(i) $l_0=8, l_1=4$ (our base model), (ii)$l_0=4, l_1=8$, (iii)$l_0=12, l_1=0$ (without the second group of transformer layers, $\{\mathbf{h}_i\}$ are used for token representations). Results are shown in Table~\ref{tab:layer}. 
Our base model (i) has better performance than setting (ii) (+0.3 in NER, +0.7 in RE).
Training setting (iii) is equal to a traditional BERT model with additional entity extraction and entity linking tasks.
The comparison with (i) and (iii) indicates that text-entity representations have better performances than text-only representations (+0.5 in NER, +0.9 in RE) in the same amount of parameters.
% \textcolor{red}{Analysis probe}

\begin{table}[t]
\centering
\begin{tabular}{lc|cc}
\hline
&\multirow{2}*{\shortstack{$l_0=8$\\$l_1=4$}}
&\multirow{2}*{\shortstack{$l_0=4$\\$l_1=8$}}
&\multirow{2}*{\shortstack{$l_0=12$\\$l_1=0$}}\\
&&&\\
\hline
BC5chem&\textbf{93.3}&93.1&93.2\\
BC5dis&\textbf{86.1}&85.7&86.0\\
NCBI&\textbf{89.1}&88.5&88.4\\
BC2GM&85.1&84.8&\textbf{86.8}\\
JNLPBA&\textbf{82.0}&81.7&78.8\\
\hline
NER&\textbf{87.1}&86.8&86.6\\
\hline
ChemProt&77.5&\textbf{77.7}&77.6\\
DDI&\textbf{81.9}&81.0&80.1\\
GAD&\textbf{84.3}&82.9&83.2\\
\hline
RE&\textbf{81.2}&80.5&80.3\\
% \hline
% Type 1&6.23&5.14&\textbf{6.91}\\
% Type 2&\textbf{0.59}&0.45&0.52\\
% Type 1&14.01&13.13&18.10\\
% Type 2&1.48&1.34&1.71\\
% \hline
% Overall&\textbf{1.44}&1.06&1.37\\
% Overall&3.26&2.90&4.06\\
\hline
\end{tabular}
\caption{Ablation studies for transformer layers count in KeBioLM on the BLURB benchmark.}
\label{tab:layer}
\end{table}

\subsection{UMLS Knowledge Probing}
We establish a probing dataset based on UMLS triplets to evaluate how LMs understand medical knowledge via pretraining.

\subsubsection{Probing Dataset}
UMLS triplets are stored in the form of $(s,r,o)$ where $s$ and $o$ are CUIs in UMLS and $r$ is a relation type.
We generate two queries for one triplet based on names of CUIs and relation type:
\begin{itemize}
    \item $Q_1$: \textit{[CLS]} $s$ $r$ \textit{[MASK] [SEP]}
    \item $Q_2$: \textit{[CLS] [MASK]} $r$ $o$ \textit{[SEP]}
\end{itemize}
For example, we sample a triplet and terms of corresponded entities (\textit{C0048038}:apraclonidine, \textit{may\_prevent}, \textit{C0028840}:ocular hypertension).
We remove the underscores of relation names and generate two queries (we omit \textit{[CLS]} and \textit{[SEP]}):
\begin{itemize}
    \item $Q_1$: apraclonidine may prevent \textit{[MASK]}.
    \item $Q_2$: \textit{[MASK]} may prevent ocular hypertension.
\end{itemize}
For relation names end with ``of'', ``as'' , and ``by'', we add ``is'' in front of relation names.
For instance, \textit{translation\_of} is converted to \textit{is translation of}, \textit{classified\_as} is converted to \textit{is classified as}, and \textit{used\_by} is converted to \textit{is used by}.
Commonly, different relation triplets can generate same query since triplets may overlap $(s,r,-)$ or $(-,r,o)$ with each other.
We deduplicate all repeat queries and randomly choose at most 200 queries from all relation types in UMLS.
After deduplication, one query can have multiple CUIs as answers.
For example:
\begin{itemize}
    \item $Q$: \textit{[MASK]} may treat essential tremor.
    \item $A_1$: \textit{C0282321}: propranolol hydrochloride
    \item $A_2$: \textit{C0033497}: propranolol
\end{itemize}
We summarize our generated UMLS relation probing dataset in Table~\ref{tab:umls}.
Unlike LAMA \cite{petroni-etal-2019-language} and X-FACTR \cite{jiang-etal-2020-x} that contain less than 50 kinds of relation, our probing task is a more difficult task requiring a model to decode entities over 900 kinds of relations.
% While KeBioLM achieves the highest macro-recall@5, it gets lower scores compared with LAMA \cite{petroni-etal-2019-language} and X-FACTR \cite{jiang-etal-2020-x}.
% Their datasets contain less than 50 kinds of relation, while UMLS has over 900 kinds of relation.
% Multi-token decoding further increases the probing difficulty of our dataset.
% We generate 143,771 pseudo sentences from 922 kinds of relationships in total.

\begin{table}
\centering
\begin{tabular}{ccc}
\hline
$\#$Queries&$\#$Relations&$\#$Avg. CUIs\\
\hline
143,771&922&2.39\\
\hline
\end{tabular}
\caption{The number of generated UMLS relation probing dataset.}
\label{tab:umls}
\end{table}

\subsubsection{Multi [MASK] Decoding}
To probe PLMs using generated queries, we require models to recover the masked tokens.
Since biomedical entities are usually formed by multiple words and each word can be tokenized into several wordpieces \cite{wu2016google}, models have to recover multiple \textit{[MASK]} tokens.
We limit the max length of one entity is 10 for decoding.

We decode the multi \textit{[MASK]} tokens using the confidence-based method described in \citet{jiang-etal-2020-x}.
We also implement a beam search for decoding.
Unlike beam search in machine translation that decodes tokens from left to right, we decode tokens in an arbitrary order.
For each step, we calculate the probabilities of all undecoded masked tokens based on original input and decoded tokens.
We predict only one token within undecoded tokens with the top $B=5$ accumulated log probabilities.
Decoding will be accomplished after count of \textit{[MASK]} times iterations and we keep the best $B=5$ decoding results.
We skip the refinement stage since it is time-consuming and does not significantly improve the results.

\subsubsection{Evaluation Metric}

Since multiple correct CUIs exist for one query, we consider a model answering the query correctly if any decoded tokens in any \textit{[MASK]} length hit any of the correct CUIs.
We evaluate the probing results by the relation-level macro-recall@5.

\subsubsection{Probing Results}

We classify probing queries into two types based on their difficulties.
Type 1: \textbf{answers within queries} (24,260 queries);
Type 2: \textbf{answers not in queries} (119,511 queries).
Here are examples of Type 1 ($Q_1$ and $A_1$) and Type 2 ($Q_2$ and $A_2$) queries:
\begin{itemize}
    \item $Q_1$: \textit{[MASK]} has form tacrolimus monohydrate.
    \item $A_1$: \textit{C0085149}: tacrolimus
    \item $Q_2$: cosyntropin may diagnose \textit{[MASK]}.
    \item $A_2$: \textit{C0001614}: adrenal cortex disease
\end{itemize}

Table~\ref{tab:probe} summarizes the probing results of different PLMs according to query types.
Checkpoints of BioBERT and PubMedBERT miss a cls/predictions layer and cannot perform the probe directly.
% There are obvious score gaps between BlueBERT, ClinicalBERT with SciBERT and our approach. 
Compared to other PLMs, KeBioLM achieves the best scores in both two types and obviously outperforms BlueBERT and ClincalBERT with a large margin, which indicates that KeBioLM learns more medical knowledge.
% Compared to other LMs, KeBioLM achieves the best scores in both two types which indicates it learns more medical knowledge.
% All LMs achieve lower scores on type 2 sentences since they require more knowledge.

\begin{table}
\centering
\begin{tabular}{lrrr}
\hline
&Type 1&Type 2&Overall\\
\hline
% SciBERT&4.85&0.24&0.89\\
SciBERT&13.92&1.01&2.75\\
% ClinicalBERT&1.51&0.06&0.25\\
ClinicalBERT&4.19&0.33&0.79\\
% BlueBERT&1.64&0.11&0.33\\
BlueBERT&4.67&0.39&1.02\\
\hline
% KeBioLM&\textbf{6.23}&\textbf{0.59}&\textbf{1.44}\\
KeBioLM&\textbf{14.01}&\textbf{1.48}&\textbf{3.26}\\
\hline
\end{tabular}
\caption{Results of the probing test in terms of Recall@5.}
\label{tab:probe}
\end{table}

Table~\ref{tab:probe_example} lists some probing examples.
SciBERT can decode medical entities for \textit{[MASK]} tokens which may be unrelated.
KeBioLM decodes relation correctly and is aware of the synonyms of hepatic.
KeBioLM states that \textit{Vaccination may prevent tetanus} which is a correct but not precise statement.

\begin{table*}[ht]
\centering
\begin{tabular}{c|ccc}
\hline
Query \& Answer CUI&SciBERT&KeBioLM\\
\hline
omalizumab may treat \textit{[MASK]}&migraine&\textbf{asthma}\\
\textit{C0004096}: asthma&the disease&severe allergic asthma\\
\hline
phentolamine may diagnose \textit{[MASK]} &depression&\textbf{pheochromocytoma}\\
\textit{C0031511}: phaeochromocytoma&the serotonin syndrome&renovascular hypertension\\
\hline
\textit{[MASK]} is noun form of hepatic&it&\textbf{liver}\\
\textit{C0023884}: liver&the form of hepatic&hepatic only\\
\hline
\textit{[MASK]} may prevent tetanus&it&vaccination\\
\textit{C0305062}: tetanus toxoid&bcg vaccination&prophylactic tetanus vaccination\\
\hline
\end{tabular}
\caption{Probing examples of UMLS relation triplets. Queries and answer CUIs are listed. We only list one correct CUI for each query. For each model, one \textit{[MASK]} token decoding result and an example of multi \textit{[MASK]} decoding result are displayed. Bold text represents a term of the answer CUI.}
\label{tab:probe_example}
\end{table*}

\section{Conclusions}
In this paper, we propose to improve biomedical pretrained language models with knowledge.
We propose KeBioLM which applies text-only encoding and text-entity fusion encoding and
has two additional entity-related pretraining tasks: entity detection and entity linking.
Extensive experiments have shown that KeBioLM outperforms other PLMs on NER and RE datasets of the BLURB benchmark.
We further probe biomedical PLMs by querying UMLS relation triplets, which indicates KeBioLM absorbs more biomedical knowledge than others.
% Compared with other knowledge probing tasks, biomedical LMs still have room for improvement.
%We integrate the relation triplets knowledge to LMs by initializing entity embeddings using TransE.
%Modeling relation information in LMs directly may enhance the text and entity representations and we leave this to further studies.
In this work, we only leverage the relation information in TransE to initialize the entity embeddings. We will further investigate how to directly incorporate the relation information into LMs in the future.

\section*{Acknowledgements}
We would like to thank the anonymous reviewers for their helpful comments and suggestions.
This work is supported by Alibaba Group through Alibaba Research Intern Program.

\bibliography{anthology,custom}
\bibliographystyle{acl_natbib}

\end{document}